\documentclass{article}

    \PassOptionsToPackage{numbers}{natbib}
\usepackage[preprint]{neurips_data_2024}

\usepackage[utf8]{inputenc} % allow utf-8 input
\usepackage[T1]{fontenc}    % use 8-bit T1 fonts
\usepackage{hyperref}       % hyperlinks
\usepackage{url}            % simple URL typesetting
\usepackage{booktabs}       % professional-quality tables
\usepackage{amsfonts}       % blackboard math symbols
\usepackage{nicefrac}       % compact symbols for 1/2, etc.
\usepackage{microtype}      % microtypography
\usepackage{xcolor}         % colors
\usepackage{mathtools}
\usepackage{tabularx}
\usepackage{graphicx}
\usepackage{subcaption}

\title{RAGBench: Explainable Benchmark for Retrieval-Augmented Generation Systems}

\author{
  Robert Friel$^*$ \\
  Galileo Technologies Inc.\\
  \texttt{rob@rungalileo.io} \\
  \And
  Masha Belyi$^*$ \\
  Galileo Technologies Inc.\\
  \texttt{masha@rungalileo.io} \\
  \And % use /AND for new line
  Atindriyo Sanyal \\
  Galileo Technologies Inc.\\
  \texttt{atin@rungalileo.io} \\
}

\begin{document}

\maketitle
\def\thefootnote{*}\footnotetext{Equal Contributions}\def\thefootnote{\arabic{footnote}}

\begin{abstract}
Retrieval-Augmented Generation (RAG) has become a standard architectural pattern for incorporating domain-specific knowledge into user-facing chat applications powered by Large Language Models (LLMs). RAG systems are characterized by (1) a document retriever that queries a domain-specific corpus for context information relevant to an input query, and (2) an LLM that generates a response based on the provided query and context. However, comprehensive evaluation of RAG systems remains a challenge due to the lack of unified evaluation criteria and annotated datasets. In response, we introduce \texttt{RAGBench}: the first comprehensive, large-scale RAG benchmark dataset of 100k examples. It covers five unique industry-specific domains and various RAG task types. \texttt{RAGBench} examples are sourced from industry corpora such as user manuals, making it particularly relevant for industry applications. Further, we formalize the TRACe evaluation framework: a set of explainable and actionable RAG evaluation metrics applicable across all RAG domains. We release the labeled dataset at \url{https://huggingface.co/datasets/rungalileo/ragbench}. \texttt{RAGBench} explainable labels facilitate holistic evaluation of RAG systems, enabling actionable feedback for continuous improvement of production applications. Thorough extensive benchmarking, we find that LLM-based RAG evaluation methods struggle to compete with a finetuned RoBERTa model on the RAG evaluation task. We identify areas where existing approaches fall short and propose the adoption of RAGBench with TRACe towards advancing the state of RAG evaluation systems.

% We evaluate existing prompt-based RAG evaluation approaches on the new benchmark, as well as fine-tune a DeBERTA-large encoder to predict \texttt{RAGBench} labels.

% We demonstrate that DeBERTA-large outperforms general prompt-based LLM judges on the RAG evaluation task.

% Through extensive analysis, we outline drawbacks of existing RAG evaluation approaches and suggest future directions for improvement. 
\end{abstract}

\section{Introduction}
% \citet{adlakha2023evaluating} evaluate LLMS.
Despite remarkable reasoning and conversational abilities, out-of-the-box pre-trained Large Language Models (LLMs) struggle to reason about out-of-domain, knowledge-intensive queries \citep{lewis2020rag, huang2024survey}. In response, Retriever-Augmented Generation (RAG) systems \citep{lewis2020rag, lee-etal-2019-latent} are becoming increasingly popular in user-facing dialogue applications \citep{siriwardhana-etal-2023-rag-for-odqa}. Generally, RAG systems comprise a retriever component that queries relevant documents from an in-domain corpus and a downstream LLM generator model that incorporates the retrieved documents along with the original user query to output an informed response. The additional context helps ground the LLM in factual information and has been shown to boost performance on knowledge-intensive tasks \citep{lewis2020rag}.

% \citep{liu-etal-2023-evaluating} find that generative search engine responses often contain unsupported facts.
  
\begin{figure}
  \centering
  \includegraphics[width=\textwidth]{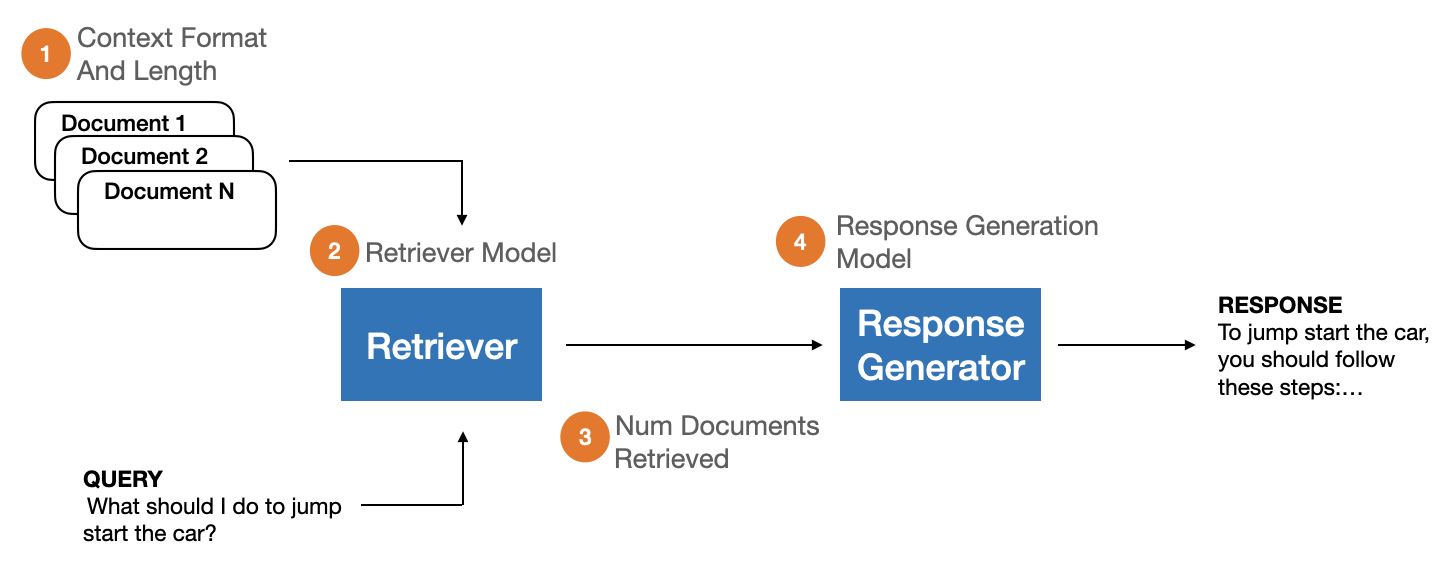}
  \caption{RAG system workflow, with highlighted variable parameters: (1) Context format and length, (2) retriever model, (3) number of retrieved documents, and (4) generation model.}
  \label{figure-rag-workflow}
\end{figure}

Still, when used in production settings, RAG systems are prone to hallucinations as the generator model struggles to retrieve relevant information from the context \citep{adlakha2023evaluating, rashkin-etal-2021-increasing, chiesurin-etal-2023-dangers}. In the absence of a one-fits-all approach, application-specific RAG systems must be fine-tuned for optimal performance on domain-specific tasks. However, the choice of retriever and generator models for each application is complex and has serious implications on overall system quality and costs. With numerous commercial and open-source generative LLMs readily available\footnote{\url{https://huggingface.co/spaces/lmsys/chatbot-arena-leaderboard}} and many variable parameters in the RAG system design (Figure \ref{figure-rag-workflow}), tuning an optimal system for a particular RAG application involves iterative evaluation of multiple configurations. This motivates the need for automated RAG evaluation solutions.

% Further, post-deployment RAG in production is challenging and not addressed by any existing solutions.

In response, automated RAG evaluation systems like RAGAS \citep{es-etal-2024-ragas} and TruLens \citep{trulens} have emerged. These systems adopt a zero-shot LLM prompt-based approach to predict a set of curated RAG evaluation metrics. However, the lack of unified RAG benchmarks makes it difficult to compare approaches against each other. Each new study designs a new dataset, often employing LLMs as generators and labelers \citep{es-etal-2024-ragas, saadfalcon2024ares, chen2023rgb}, which renders them irreproducible. A few benchmarks like RGB \citep{chen2023rgb}, AttributionBench \citep{attributionbench} and RAGTruth \citep{wu2023ragtruth} have been proposed recently, but they are small in size and target a disjoint set of labels. The exact RAG evaluation criteria also vary from study to study. ARES \citep{saadfalcon2024ares} and RAGAS \citep{es-etal-2024-ragas} define a \textit{context relevance} metric to evaluate the quality of the retrieved documents, along with \textit{answer relevance} and \textit{faithfulness} to evaluate the quality of the generative model. However, others have explored other metrics like \textit{correctness} \citep{adlakha2023evaluating} \textit{noise rejection} and \textit{robustness} \citep{chen2023rgb}, to name a few. Finally, most studies evaluate on small in-domain evaluation datasets that are specific to each new application \citep{saadfalcon2024ares, sadat2024delucionqa, es-etal-2024-ragas, adlakha2023evaluating, chen2023rgb}, leaving cross-domain generalization an open question.

% Can cite other LLM eval benchmarks that are not RAG specific. While a number of benchmarks for hallucintaion detection and general LLM evaluation exist, there is nothing specific to RAG.

% Evaluation format also varies: sometimes takes the form of direct assessment (attrbench), and sometimes pairwise ranking (ragas, ares).

In this work we propose RAGBench: a comprehensive dataset for training and benchmarking RAG evaluation models. RAGBench comprises data sourced from multiple domains along with a comprehensive suite of evaluation metrics. Specifically, we adopt existing metric definitions for \textit{context relevance}, \textit{answer faithfulness} \citep{es-etal-2024-ragas, saadfalcon2024ares} and introduce two new metrics: \textit{context utilization} and \textit{answer completeness}. We argue that this new suite of metrics better describes the overall RAG system performance, with the potential to provide granular, actionable insights to the RAG practitioner.

We evaluate state-of-the art LLMs and existing RAG evaluation systems on RAGBench. We find that a 400M-parameter DeBERTa-large model that was fine-tuned on RAGBench outperforms few-shot LLM judges across numerous domains and task types. We highlight this result to motivate future work aimed at leveraging these data for advancing RAG evaluation models and improving on the proposed benchmark.

% [Identify any difficult domains, if any]. We find that a fine-tuned DeBERTa model performs similar to a GPT-3.5 and is a reliable, cost-effective way to monitor these metrics at scale in production settings. However, more work is required to match GPT-4 performance.  [Add any other main conclusions]. 

% We summarize our contributions as follows:
% \begin{itemize}
%     \item \textbf{Release RAGBench}: a comprehensive benchmark dataset for training and benchmarking RAG evaluation models for real-world applications. With the release of RAGBench, we enable the development of foundational, cross-domain, RAG evaluation models.
%     \item Formulate the \textbf{TRAce evaluation framework}: new explainable RAG evaluation metric formulations for \textit{context relevance}, \textit{context utilization}, \textit{answer completeness}, and \textit{answer adherence}.
%     \item Identify gaps in LLM-based RAG evaluation approaches.
%     Demonstrate that a fine-tuned DeBERTA-large encoder attains competitive performance on RAGBench. The model, trained on RAGBench data, is finetuned for cross-domain generalizability. We propose this as a foundational RAG evaluation model benchmark that can be further fine-tuned for additional metrics or special RAG use cases.
% \end{itemize}

\section{Related Work}
% \paragraph {RAG benchmarks} Numerous general LLM evaluation benchmarks, such as ChatbotArena \citep{zheng2023judging} have been proposed in past work. However, human preference datasets, constructed through pairwise comparisons, have limitations. While these data are appropriate for fine-tuning general purpose LLM judges, they are insufficient for building RAG evaluation systems because preference judgements under-represent important RAG dimensions like factuality and completeness of the response \citep{hosking2024human}.

We differentiate our work from existing ground-truth RAG datasets like ChatRAGBench \citep{liu2024chatqa}, CRAG \citep{yang2024cragcomprehensiverag}, DomainRAG \citep{wang2024domainragchinesebenchmarkevaluating}, and ALCE \citep{gao-etal-2023-alce}. These datasets contain RAG samples with ground-truth responses and are used for end-to-end evaluation of RAG systems via response-level metrics like exact match or ROUGE scores. In contrast, we design RAGBench to enable development of more mature evaluation systems that effectively evaluate different parts of the RAG system along multiple dimensions like retriever relevance, adherence and completeness of the response.

% but lacks the granular component-specific labels that we release with RAGBench}

% \textsc{ChatRAGBench} \citep{liu2024chatqa} is a recent initiative that is similar in intent to our work in that it contributes a large-scale unified RAG benchmark. However, \textsc{ChatRAGBench} only contains ground truth responses and lacks the granular component-specific labels that we release with RAGBench. 
% \textcolor{blue}{Similar:, CRAG, DomainRAG. While these datasets may be used for end-to-end RAG system evaluation, they are not suitable for training and benchmarking RAG evaluation models. the evaluation is limited to response-level metrics, and lacks the granular component-specific labels that we release with RAGBench. }
% As future work, we can consider annotating \textsc{ChatRAGBench} with the schema proposed in this paper, to further scale RAGBench.

Various RAG benchmarks focus specifically on hallucination detection \citep{wu2023ragtruth, sadat2024delucionqa, li-etal-2023-halueval, chen2023felm}.
FELM \citep{chen2023felm} is a multi-domain and task dataset with factuality labels for open domain QA. RAGTruth \citep{wu2023ragtruth}, DelucionQA \citep{sadat2024delucionqa}, and HaluEval \citep{li-etal-2023-halueval} are RAG-specific datasets with both synthetic as well as human-annotated labels for hallucinations in LLM reseponses. While these are appropriate benchmarks for hallucination detection models, they do not offer the level of granularity we offer with RAGBench that is necessary to understand the RAG system as a whole.

% RAGTruth \citep{wu2023ragtruth} is another recent effort at a RAG Benchmark. RAGTruth combines QA, Data-toText, and Summarization RAG data with human annotated hallucinated spans in the response. While it is an excellent benchmark for hallucination detection, it does not offer the level of granularity we offer with RAGBench that is necessary to understand the RAG system as a whole.

\paragraph{RAG evaluation} Recently, several parallel efforts have proposed approaches to automated RAG evaluation. In RAGAS \citep{es-etal-2024-ragas}, the authors query an LLM-judge (GPT-3.5) with a curated prompt to evaluate \textit{context relevance}, \textit{answer relevance} and \textit{faithfulness} of a RAG response. Next, \citet{saadfalcon2024ares} propose ARES, a framework for fine-tuning smaller NLI models to predict the same metrics. In parallel, \citet{chen2023rgb} develop a heuristic system to probe LLM's robustness to noisy and irrelevant context documents, and \citet{adlakha2023evaluating} explore heuristic algorithms to estimate RAG \textit{correctness} and \textit{faithfulness}. The lack of established RAG benchmarks makes it difficult to compare these approaches against each other. We aim to address this limitation by introducing RAGBench.

% - Faithfulness is closely related to groundedness.

\paragraph{Finetuned RAG evaluation models} Fine-tuned LLM judges are another a common way to approach the LLM evaluation task \citep{kim2024prometheus, yue-etal-2023-automatic, wu2023ragtruth}. A number of studies also leverage small, fine-tuned Natural Language Inference (NLI) models for RAG hallucination detection \citep{bohnet2023attributed, attributionbench, saadfalcon2024ares}. NLI models measure the degree of entailment between a premise and a hypothesis, which has been successfully repurposed for evaluating LLM response attribution in RAG setting. In this work, we train and evaluate an NLI model for RAG evaluation using RAGBench. The fine-tuned model not only outperforms LLM judges in hallucination/attribution detection but also excels on the new RAG evaluation metrics we propose.

% A few studies also propose tuning small Natural Language Inference (NLI) models to to evaluate LLM output . Larger fine-tuned LLM judges are also a common benchmar (RAGTruth, find others in LUNa refs). NLI is closely related to the RAG hallucination task task (explain more).

% However, existing work focuses on select metrics \citep{attributionbench} or application-specific models \citep{saadfalcon2024ares}. Lack of unified datasets makes it difficult to access how well these models to across different domains and RAG task formulations. In this work we demonstrate that a fine-tuned DeBERTA NLI model is competitive with zero-shot LLM evaluators across all RAGBench domains and can be a competitive foundational RAG evaluation model.

% ARES demonstrates generalizability across 2 RAG domains, but falls short of proposing a foundational cross-domain model.

% A lot of previous work focusing on evaluating LLM Attribution (CITE)\citep{yue2023automatic} . Recently some authors have proposed custom RAG-specific metrics:
% \begin{itemize}
%     \item RAGAS \citep{es-etal-2024-ragas}: context and answer relevance and faithfulness(adherence)
%     \item ARES \citep{saadfalcon2024ares}: context and answer relevance and faithfulness(adherence)
%     \item \citet{chen2023rgb} (Chen at al, 2023): RGB - noise robustness, negative rejection, information integration, counterfactual robustness. 
%     \item (Adlakha et al, 2023) Instruct-QA: correctness, faithfulness. Faithfulness is similar to our completeness. 
% \end{itemize}

\section{RAGBench Construction}
\label{section:ragbench-construction}
% We first define the evaluation metrics in \ref{metric_definitions}, then describe the process of dataset curation in Sections \ref{section:llm-annotator}-\ref{dataset_construction}

\subsection{Component Datasets}
\label{dataset_construction}
RAGBench is a collection of real-world datasets that span different domains and RAG task types. We source data from open-book Question-Answer (QA) datasets (CovidQA \citep{moller-etal-2020-covidqa}, PubmedQA \citep{jin-etal-2019-pubmedqa}, HotpotQA \citep{yang2018hotpotqa}, MS Marco \citep{nguyen2016msmacro}, CUAD \citep{hendrycks2021cuad}, EManual \citep{nandy-etal-2021-emanual}, TechQA \citep{castelli-etal-2020-techqa}, FinQA \citep{chen-etal-2021-finqa}, TAT-QA \citep{zhu-etal-2021-tatqa}, ExpertQA \citep{malaviya2024expertqa}, HAGRID \citep{kamalloo2023hagrid}), as well one that was specifically adapted for RAG (DelucionQA \citep{sadat2024delucionqa}). We transform all 12 component datasets to a standardized RAG format with consistent annotations. To best represent real-world RAG scenarios, we vary a number parameters to construct the benchmark: the source domain, number of context documents, context token length, and the response generator model Figure \ref{figure-rag-workflow} illustrates where these variable parameters fall in the RAG pipeline.

\paragraph{Source Domains} RAGBench comprises five distinct domains: bio-medical research (PubmedQA, CovidQA), general knowledge (HotpotQA, MS Marco, HAGRID, ExperQA), legal contracts (CuAD), customer support (DelucionQA, EManual, TechQA), and finance (FinBench, TAT-QA). We select these specific domains based on availability of data, and applicability to real-world RAG applications across different industry verticals. For detailed descriptions of each component data source, refer to Appendix \ref{appendix-component-datasets}.

\paragraph{Context Token Length} Context token length in RAGBench ranges from 100 to 11k tokens, which we report in Table \ref{table:component-datasets}. Notably, CUAD documents feature long contexts of up to 11k tokens each, compared to the relatively short context in PubMedQA.

% \paragraph{Format} Product Manuals - straightforward information extraction. HotpotQA requires logical reasoning across multiple document to derive the correct answer. FinBench requires numerical reasoning from structured tables.

\newcolumntype{s}{>{\hsize=.3\hsize}X}
\begingroup
\begin{table}
  \caption{RAGBench component datasets.}
  \label{table:component-datasets}
  \centering
  \small
  \begin{tabularx}{\textwidth}{lXXXsssss}
    \toprule
    % \multicolumn{2}{c}{Part}                   \\
    % \cmidrule(r){1-2}
    Dataset & Domain & Document Source & Question Source & \#docs & doc length & \#Train & \#Dev & \#Test \\
    \midrule
    \textbf{PubMedQA} & biomedical \newline research & research \newline abstracts & automated heuristics & 4 & 99 & 19.5k & 2.5k & 2.5k \\
    % \hline
    \textbf{CovidQA-RAG}& biomedical \newline research & research \newline papers & expert & 4 & 122 & 2.5k & 534 & 492 \\
    % \midrule
    \textbf{HotpotQA} & general \newline knowledge & wikipedia & crowd-sourced & 4 & 126 & 3.7k & 847 & 776 \\
    \textbf{MS Marco} & general \newline knowledge & web pages & user \newline web queries & 10 & 94 & 3.7k & 790 & 839 \\
    \textbf{HAGRID} & general knowledge & wikipedia & expert & 3 & 153 & 2.0k & 322 & 1.3k \\
    \textbf{ExpertQA} & general knowledge & google search & expert & 3 & 548 & 1.6k & 202 & 203 \\
    % \midrule
    \textbf{CUAD}& legal & legal \newline contracts & expert & 1 & 11k & 1.5k & 506 & 508 \\
    % \midrule
    \textbf{DelucionQA}& customer\newline support & Jeep manual & LLM & 3 & 296 & 1.5k & 177 & 182 \\
    \textbf{EManual}& customer\newline support & TV manual & annotator & 3 & 165 & 1k & 132 & 132 \\
    \textbf{TechQA}& customer\newline support & Technotes & tech forums & 5 & 1.8k & 1.2k & 302 & 310 \\
    % \midrule
    \textbf{FinQA}& finance & earning \newline reports & expert & 3 & 310 & 12k & 1.7k & 2.2k \\
    \textbf{TAT-QA}& finance & financial \newline reports & expert & 5 & 96 & 26k & 3.2k & 3.2k \\
    \midrule
    \textbf{Total} &&&&&& 78k & 12k & 11k \\
    \bottomrule
  \end{tabularx}
\end{table}
\endgroup

\paragraph{Task Types} We curate RAGBench to inlcude a variety of difficult RAG task types. Customer support datasets simulate a common application of RAG in industry settings. FinQA and TAT-QA require numerical reasoning over hybrid tabular and text data. HotpotQA, CovidQA, and PubMedQA necessitate retrieval and reasoning over multiple context docs. The CUAD dataset is a challenging addition to RAGBench for several reasons: (i) it represents a difficult and highly-specialized real-world domain in which of-the-shelf pre-trained LLM models struggle to perform well \citep{magesh2024hallucinationfree}, and (ii) it is equally challenging in RAG context due to very long context lengths of legal contract documents.

\paragraph{Question Sources} All component datasets include domain-specific questions that represent real-world user queries about various topics. Questions for DelucionQA, HotpotQA, and EManual are crowd-sourced; questions for CovidQA, CUAD, HAGRID, ExpertQA, and FinQA are composed by domain experts; MS Marco is sourced from real-world user web search queries; likewise, TechQA questions are user queries posted on IBM technical forums; PubMedQA is the only dataset with automatically-generated questions from research article titles. 

% \paragraph{Context Retrieval} For CuAD, HotpotQA, and MS Marco, we leverage the context and questions provided with the original dataset. CuAD has one contract per question, docs are long context length. HotpotQA - multi-hop reasoning across multiple Wikipedia extracts; factual questions. For all other component datasets, we implement dense retrieval with FAISS to query a between 3-10 context documents per question.

\paragraph{Response Generation} For each component dataset we generate responses with LLMs. Exceptions to this are HAGRID and ExpertQA datasets, which contain LLM-generated responses in the original data. To introduce variability into the dataset, we generate two responses per input with different modes: GPT-3.5 (gpt-3.5-0125) and Claude 3 Haiku. Both are proprietary models that are offered at a reasonable price point\footnote{https://openai.com/api/pricing/, https://www.anthropic.com/api}, which we believe make them suitable candidates for generating real-world RAG responses. For CUAD we only generate responses with Claude 3 Haiku due to prohibitively long context lengths that exceed the GPT-3.5 16k token limit. To encourage a diverse distribution of labels in RAGBench, we use a basic prompt (Appendix \ref{appendix-response-generation-prompt}) that does not explicitly require the model to stick to the provided context when generating the response. We set the temperature to 1.0 for generation.

% We find that the LLM is more likely to hallucinate in this case, leading to a varied distribution of labels in RAGBench.

\paragraph{Data Splits} We split each component dataset into train, validation, and test sets, ensuring there is no overlap in queries across splits from the same data source. RAGBench totals 100k samples, split across train, validation, and test sets. Component dataset statistics are reported in Table \ref{table:component-datasets}.
% \subsection{Dataset Analytics}

\subsection{TRACe Evaluation Framework}
\label{metric_definitions}

We propose a suite of four comprehensive metrics to evaluate the quality of the retriever and the response generator components of RAG. An optimal RAG system must balance accuracy and efficiency. The retriever should precisely return all the necessary information to address the user query, avoiding any superfluous data. The generator must effectively utilize the retrieved information, ensuring the response is strictly based on the provided context without introducing any hallucinations in the output.

Towards comprehensive evaluation of the abovementioned criteria, we introduce the TRACe evaluation framework to measure u\textbf{T}ilization, \textbf{R}elevance, \textbf{A}dherence, and \textbf{C}ompleteness of a RAG system. Utilization, Adherence, and Completeness measure the quality of the generator. Adherence here is synonymous with previously proposed \textit{answer faithfullness}, \textit{groundednes}, and \textit{attribution}, all terms used in literature to measure how well an LLM output adheres to a source of factual information. Relevance measures the quality of the retriever output with respect to the query. Below we formalize the definition of each metric.

\begin{figure}
  \centering
  \includegraphics[width=\textwidth]{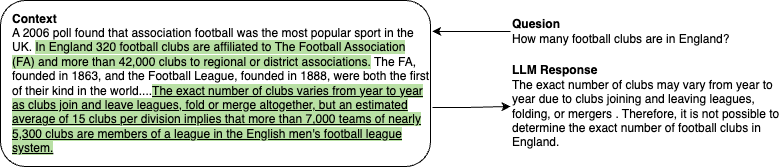}
  \caption{Example of RAG Question, Context, and Response. Relevant context spans are highlighted, and utilized spans are \underline{underlined}.}
  \label{figure:rel-util-spans}
\end{figure}

\paragraph{Definitions} Let $D$ be a set of context documents $\{d_1...d_n\}$ retrieved for a RAG input query. We define a set of \textit{relevant} tokens in $d_i$ as $R_i=\{t_1, ... t_r\}$. $R_i$ encodes information in context document $d_i$ that is useful for answering the query. Similarly, we define $U_i=\{t_1, ... t_u\}$ as the set of \textit{utilized} tokens in document $d_i$, which reflect information that the generation model is using to produce a response. Refer to Figure \ref{figure:rel-util-spans} for a visual representation of \textit{relevant} and \textit{utilized} spans. $Len(x)$ measures the length of strings in $x$, which can be interpreted as character length, token length, or sentence length. For calculating ground-truth metrics, we employ sentence-length, since it aligns best with our annotation schema (Section \ref{section:llm-annotator}). However, token or character length may also be suitable for other use cases.

% 
% (\textcolor{red}{Add figure with highlighted relevant and utilized strings})

\paragraph{Context Relevance} Context Relevance is defined in \citep{es-etal-2024-ragas, saadfalcon2024ares} as the fraction of the retrieved context that is relevant to the input query. Low relevance points to an inefficient retriever that supplies excess information to the generation model. Long context inputs into the generator may accrue unnecessary costs, as well as compromise the quality of the generated output. We measure relevance of context document $d_i$ as:
\begin{equation}
    \text{document relevance} = \frac{{Len(R_i)}}{Len(d_i)}
\end{equation}
Example-level relevance can be aggregated over all context documents in the example as:
\begin{equation}
\label{eq:example-relevance}
    \text{example relevance} = \frac{\sum_{i=1}^{|D|} {Len(R_i)}}{\sum_{i=1}^{|D|} Len(d_i)}
\end{equation}

\paragraph{Context Utilization} Context Utilization is a new metric introduced in TRACe. We aim to measure the the fraction of the retrieved context that is used by the generator to produce the response.
% Utilization is closely related to attribution \citet{rashkin2023measuringAttributionInNLG}.
Low Utilization in combination with low Relevance points to a greedy retriever, while low Utilization alone points to a weak generator that fails to leverage the provided context efficiently. Document-level and example-level Utilization are defined as:
\begin{equation}
\label{eq:example-utilizaion}
    \text{document utilization} = \frac{{Len(U_i)}}{Len(d_i)}
    \quad%\text{and}\quad
    \text{example utilization} = \frac{\sum_{i=1}^{|D|} {Len(U_i)}}{\sum_{i=1}^{|D|} Len(d_i)}
\end{equation}

\paragraph{Completeness} Completeness is another new metrics we introduce to measure how well the response incorporates all the relevant information in the context. Note that this is different from Utilization; it is possible to have high Relevance and high Utilization, but low Completeness when the generator utilizes irrelevant information in the context to produce a low quality response. Completeness for document $d_i$ is calculated as the fraction of utilized substrings among all relevant substrings: 
\begin{equation}
\label{eq:example-completeness}
    \text{completeness} = \frac{Len(R_i \cap U_i)}{Len(R_i)}
\end{equation}
And can be extended to example-level by considering all relevant and utilized substrings across all context documents. 

\paragraph{Adherence} Adherence is designed to detect hallucinations in RAG responses. Our definition of Adherence is synonymous with answer faithfullness \citep{es-etal-2024-ragas, saadfalcon2024ares}, groundednes \citep{trulens}, and attribution \citep{rashkin2023measuringAttributionInNLG}. For alignment with existing hallucination detection approaches, we define example-level adherence as a boolean indicating whether or not all parts of the response are grounded in the context. However, in our annotation schema (Section \ref{section:llm-annotator}) we also define $A_i=\{t_1, ... t_a\}$ as the set of response tokens that are supported by the context to enable granular Adherence evaluation.

\begin{figure}
  \centering
  \includegraphics[width=\textwidth]{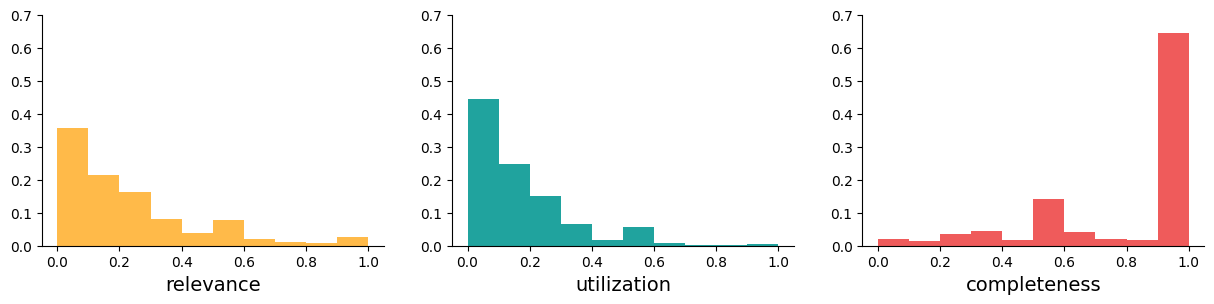}
  \caption{Distributions of relevance, utilization, and completeness labels in RAGBench. Y-axis is normalized to visualize densities.}
  \label{figure:full-ragbench-stats}
\end{figure}

\subsection{RAGBench Statistics}
RAGBench component datasets contain between 1\% - 20\% hallucinations. ExpertQA, CovidQA, and MS Marco contain the highest fraction of hallucinated responses (12\%, 16\%, and 13\%, respectively), while Cuad, FinQA, and TAT-QA contain the least (about 1\% for each). We visualize distributions of relevance, utilization, and completeness scores in Figure \ref{figure:full-ragbench-stats}.

% In combination, these 5 metrics provide a holistic evaluation of a RAG system. Describe sample scenarios: Low utilization points to either a greedy retriever or an inefficient LLM; low utilization in combination with low relevance points to an inefficient retriever but a potentially good LLM; low utilization, relevance, and adherence points to hallucinations. [Add image illustrating the metrics. highlight relevant, utilized chunks and call-out example action items, e.g. improve retriever, reduce chunk size, or improve LLM].

% Explain why we omit Answer Relevance. \textcolor{red}{TODO: Show that this metric from RAGAS is > 0.9 for 90\% of RAGBench, argue that it is not informative.}

\subsection{LLM annotator}
\label{section:llm-annotator}
% As many before us (CITE), we leverage a powerful LLM for generating high quality labels. 
We prompt GPT-4 (\texttt{gpt-4-0125-preview}) to produce ground truth Adherence, Relevance, and Utilization labels for input (\textit{documents}, \textit{query}, \textit{response}) tuples in RAGBench. Completeness is easily derived from span-level Relevance and Utilization annotations, thus we don't request explicit annotations for it.

For high quality labels, we use proven techniques like chain of thought \citep{wei-2022-chain-of-thought} that have been shown to maximize the correlation between GPT-4 and human judgements \citep{ye2024flask, zheng2023judging}.
For relevance and utilization we request the LLM-annotator to directly identify relevant and utilized sub-strings in the input documents. 
For adherence, we instruct the LLM to identify which response sentences, if any, are supported by the provided context. We can then derive an example-level boolean adherence label by checking if all response sentences are supported. The exact prompt used for annotation is provided in Appendix \ref{appendix-gpt-prompts}. We apply post-processing steps to ensure high quality, reliable annotations from our GPT-labeler, which we outline in Appendix \ref{appendix:annotation-post-processing}.

% We further validate our annotation approach in Section \ref{section:anntoation-validation}, and discuss the limitations of using an LLM-annotator in Section \ref{section:limitations}.
\begingroup
% \newcolumntype{s}{>{\hsize=.5\hsize}X}
\renewcommand{\arraystretch}{1.2} % increase vertical spacing between rows
\begin{table}
  \caption{GPT-4 annotator achieves high alignment with human judgements. We report F1 and Accuracy alignment metrics on human annotated subsets of DelucionQA. Adherence annotations are evaluated against DelucionQA human-annotated labels. Relevance and Utilization are evaluated against \textbf{DelucionQA(40)}: a subset of 40 examples randomly sampled from the DelucionQA test set and annotated by the authors.}
  \label{table:alignment-with-human-labels}
  \centering
  \small
  \begin{tabularx}{0.8\textwidth}{lccc}
    \toprule
     Test Set & Metric & F1 & Accuracy \\
    \midrule
    DelucionQA - example level & Adherence & 0.96 & 0.93 \\
    DelucionQA - span level & Adherence & 0.97 & 0.95 \\
    DelucionQA(40) - span level & Utilization & 0.92 & 0.94  \\
    DelucionQA(40) - span level & Relevance & 0.76 & 0.78 \\
    \bottomrule
  \end{tabularx}
\end{table}
\endgroup

\paragraph{Alignment with Human Judgements} We validate our metric formulations and labeling approach on a human annotated benchmark. DelucionQA \citep{sadat2024delucionqa} is a curated collection of user queries on the operation of Jeep’s 2023 Gladiator model. Natural language queries are first generated by an LLM, then reviewed and filtered by human annotators. Context documents are sourced from Jeep’s Gladiator User Manual, and responses are generated by various LLMs. Combined with example-level and span-level annotations for hallucination, DelucionQA represents a realistic distribution of real-world user queries and RAG responses. We find that our GPT annotator achieves 93\% and 95\% example- and span-level agreement with human judgements on the DelucionQA test split (Table \ref{table:alignment-with-human-labels}).

To validate relevance and utilization annotations, we also annotate a small subset of DelucionQA with granular relevance and utilization labels. We refer to this subset as DelucionQA(40) in Table \ref{table:alignment-with-human-labels}. Similar to adherence, we observe high correlation between relevance and utilization judgements from GPT-4 and humans. Details of the annotation process and additional validations are found in Appendix \ref{appendix-annotation-alignment}.
% RAGBench raw annotations contain token-level labels for utilization and relevance, which are converted to TRACe metrics using equations in Section \ref{metric_definitions}. We encourage future work on automated evaluators to predict the raw token-level labels, like relevant and utilized spans, rather than predicting the example-level scores directly which are less interpretable for the end user.

\begin{figure}
     \centering
     \begin{subfigure}[b]{0.45\textwidth}
         \centering
         \includegraphics[width=\textwidth]{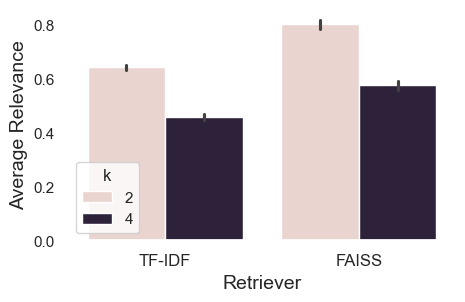}
         \caption{Retriever vs. Relevance}
         \label{fig:rag-case-study-relevance}
      \end{subfigure}
      \hfill
     \begin{subfigure}[b]{0.45\textwidth}
         \centering
         \includegraphics[width=\textwidth]{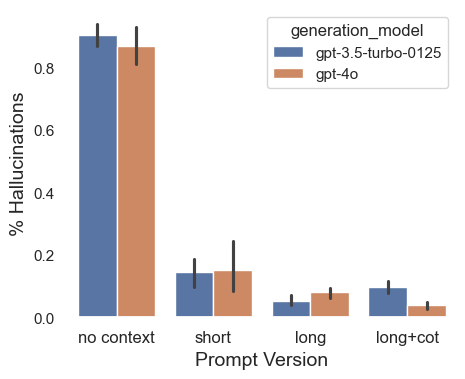}
         \caption{LLM vs. Hallucination}
         \label{fig:rag-case-study-halu}
     \end{subfigure}
     \hfill
     \begin{subfigure}[b]{0.45\textwidth}
         \centering
         \includegraphics[width=\textwidth]{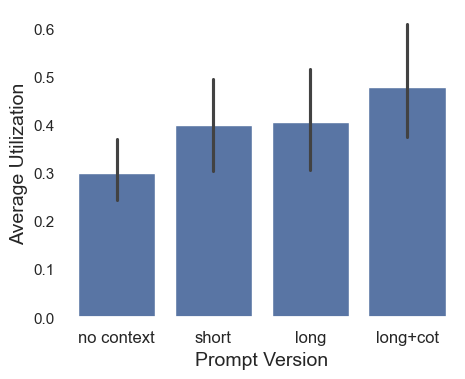}
         \caption{Prompt vs. Utilization}
         \label{fig:rag-case-study-utilization}
     \end{subfigure}
     \hfill
     \begin{subfigure}[b]{0.45\textwidth}
         \centering
         \includegraphics[width=\textwidth]{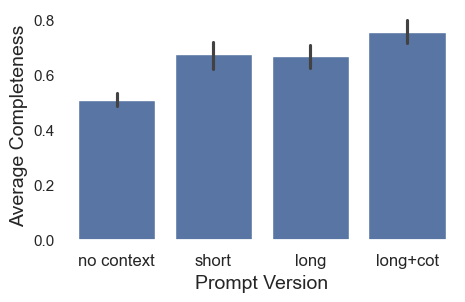}
         \caption{Prompt vs. Completeness}
         \label{fig:rag-case-study-completeness}
     \end{subfigure}
        \caption{Relationship between RAG system configuration and TRACe metrics. (a) The choice and configuration of RAG retriever component affects the average relevance of the retrieved context. In this example, a dense retriever with a low number of documents per query (k=2) yields the highest average context relevance. (b) The choice of LLM and generation prompt affect how well the RAG system utilizes the provided context. Prompting the LLM with a detailed chain-of-thought prompt leads to reduced hallucinations, as well as higher response utilization (c) and completeness (d) rates.}
        \label{fig:rag-case-study}
\end{figure}

\subsection{RAG Case Study}
We design a case study to further motivate and validate the proposed TRACe framework. We sample 100 realistic world knowledge queries from the RAGTruth \citep{wu2023ragtruth} QA training set, along with 2,512 unique context document chunks from the same dataset to use as inputs into our mock RAG systems. We simulate RAG setups of varying quality by controlling four configurable parameters: \textbf{the retriever} (sparse TF-IDF vs. dense), \textbf{number of retrieved documents} (2 vs. 4),  \textbf{generation model} (GPT-3.5-turbo vs. GPT-4o), and 4 versions of the \textbf{generation prompt template}. For illustrative purposes, we evaluate one prompt template comprising of only the question (no context) vs. three RAG-style templates that encourage the LLM to utilize the provided context and/or utilize chain-of-thought (CoT). The full generation templates are provided in Appendix \ref{appendix-rag-case-study}. We generate responses for the 100 RAGTruth queries with each of the 32 resulting RAG systems and use our LLM annotation prompt to evaluate the TRACe metrics.

Figure \ref{fig:rag-case-study} demonstrates how the different RAG configurations influence TRACe metrics. For example, we confirm that the choice of the generative LLM model affects the amount of hallucinations (or non-adherent responses) in the generated text. Not surprisingly, GPT-4o combined with a chain-of-thought prompt yields the lowest amount of hallucination compared to the weaker GPT-3.5 model and less detailed prompts. Curiously, we find that GPT-3.5 hallucinates more when prompted with CoT, which may point to limitations in the model's ability to reason about complex concepts \ref{fig:rag-case-study-halu}. Overall, we find that prompting the LLM to think step-by-step and explain its reasoning leads to higher utilization of the provided context and more complete responses (\ref{fig:rag-case-study-utilization}, \ref{fig:rag-case-study-completeness}). Finally, we show that the choice of the retriever affects average relevance of the retrieved context documents per query \ref{fig:rag-case-study-relevance}.

% Note that Possible to have high utilization and high hallucination. For a comprehensive list of possible outcomes, see Appendix Table \ref{table:rag-case-study-full-results}.

% From Prometheus Paper: Prior works employing proprietary LMs as evaluators have demonstrated not only high correlations with human evaluations but also increased speed and cost-effectiveness (Zheng et al., 2023; Liu et al., 2023b; Dubois et al., 2023; Ye et al., 2023).

% \subsection{Synthetic Data}
% We also generate synthetic RAG data using GPT-3.5. Describe the process. Put prompts and details in Appendix.

% \section{[Optional] Validate dataset and labels}
% \begin{itemize}
%     \item either through human annotation - some datasets have available annotations for relevance, attribution, and adherence, or
%     \item (probably not) using 1 source corpus, generate RAG datasets of varying quality by varying (1) the embedding model used in retriever, (2) number of docs retrieved per query, (3) the LLM generator. Then report GPT-4 metrics on these data to confirm rankings are as expected. E.g. OpenAIEmbeddings for retriever should have higher relevance than sBERT; larger models better adherence and completion.
% \end{itemize}

% \section{Alignment with Human Judgements}
% \label{section:anntoation-validation}

\section{Experiments}
\label{experiments}

\subsection{LLM Judge}
% Zero-shot LLM prompting for RAG evaluation has become standard practice (CITE).
We benchmarks a few LLM evaluators on RAGBench: (1) zero-shot GPT-3.5-judge, where we query GPT-3.5 with our annotation prompt, (2) RAGAS \citep{es-etal-2024-ragas}, and (3) TruLens \citep{trulens}. RAGAS employs a series of few-shot prompts to GPT-3.5 to measure answer groundedness (Adherence) and Context Relevance metrics. Trulens is another zero-shot prompting approach that measures answer faithfulness (Adherence) and Context Relevance.

\subsection{Fine-tuned Judge}
\label{section:deberta-training}
% We follow the approach in previous work \citep{bohnet2023attributed, gao-etal-2023-rarr, attributionbench} to leverage a pre-trained natural language inference (NLI) model as the starting point for RAG-specific fine-tuning. NLI is closely related to adherence and utilization, two labels that we are interested in predicting.

We fine-tune a DeBERTa-v3-Large \citep{he2023debertav} NLI checkpoint\footnote{https://huggingface.co/MoritzLaurer/DeBERTa-v3-large-mnli-fever-anli-ling-wanli} from \citet{laurer2022debertaNLI} with one key architecture modification: we add a shallow prediction head for each of the output RAG metrics, which allows us to compute all TRACe metrics in a single forward pass. This is both cost-effective and enables transfer learning from head to head through back-propagation down to the shared base layers. Each prediction head is a single layer feed-forward net that acts on the token-level output of the last DeBERTa layer.

We attach two heads on the context tokens to estimate Relevance and Utilization probabilities, and another head on the response tokens to estimate Adherence. For training, we broadcast sentence-level annotations to tokens, and tune to maximize token-level probabilities of Relevant, Utilized, and Adherent spans. At inference, we impose a probability threshold=0.5 to predict Relevant and Utilized spans and Adherent spans and calculate TRACe metrics using equations \ref{eq:example-relevance}, \ref{eq:example-utilizaion}, and \ref{eq:example-completeness}. For comparison with existing hallucination detection approaches, we also aggregate Adherence probabilities across the entire response to produce an example-level response adherence label. For details about training and hyperparameters, refer to Appendix \ref{appendix-deberta-training}.

%\textcolor{red}{Add small diagram of multi-head DeBRTa}.

\subsection{Evaluation}
Our granular annotation schema allows for various evaluation setups. For example, we could evaluate either span-level or example/response-level predictions. For easy comparison with existing RAG evaluation approaches that are less granular, we report area under the receiver-operator curve (AUROC) on the response-level hallucination detection task, and root mean squared error (RMSE) for example-level context Relevance and Utilization predictions.

\begingroup
% \newcolumntype{s}{>{\hsize=.5\hsize}X}
\renewcommand{\arraystretch}{1.1} % increase vertical spacing between rows
\begin{table}
  \caption{Benchmark evaluation on test splits. Reporting AUROC for predicting hallucinated responses (Hal), RMSE for predicting Context Relevance (Rel) and utilization (Util). $^*$ indicates statistical significance at 95\% confidence intervals, measured by bootstrap comparing the top and second-best results. RAGAS and Trulens do not evaluate Utilization.}
  \label{table:main-results}
  \centering
  \small
  \begin{tabularx}{\textwidth}{lsscsscsscssc}
    \toprule
    % \multicolumn{2}{c}{Part}                   \\
    % \cmidrule(r){1-2}
     & \multicolumn{3}{c}{\textbf{GPT-3.5}} & \multicolumn{3}{c}{\textbf{RAGAS}} & \multicolumn{3}{c}{\textbf{TruLens}} & \multicolumn{3}{c}{\textbf{DeBERTA}} \\
    % \cmidrule(r){2-13}
    \cmidrule(r){2-4}   \cmidrule(lr){5-7} \cmidrule(lr){8-10} \cmidrule(l){11-13}
    \textbf{Dataset} & Hal$\uparrow$ & Rel$\downarrow$ & Util$\downarrow$ 
    & Hal$\uparrow$ & Rel$\downarrow$ & Util$\downarrow$ 
    & Hal$\uparrow$ & Rel$\downarrow$ & Util$\downarrow$ 
    & Hal$\uparrow$ & Rel$\downarrow$ & Util$\downarrow$ \\
    \midrule
    \textbf{PubMedQA}       &0.51&\textbf{0.21$^*$}&\textbf{0.16}     &0.54&0.37&-     &0.62&0.45&-     &\textbf{0.80$^*$}&0.26&0.17 \\
    \textbf{CovidQA-RAG}    &0.57&0.18&\textbf{0.11}     &0.58&\textbf{0.17}&-     &0.62&0.58&-     &\textbf{0.77$^*$}&0.19&\textbf{0.11} \\
    \midrule
    \textbf{HotpotQA}       &0.59&\textbf{0.11}&\textbf{0.08}     &0.62&0.14&-     &0.64&0.73&-     &\textbf{0.85$^*$}&\textbf{0.11}&\textbf{0.08} \\
    \textbf{MS Marco}       &0.65&0.23&0.11     &0.63&0.25&-     &0.62&0.61&-     &\textbf{0.70}&\textbf{0.22}&\textbf{0.10} \\
    \textbf{HAGRID}         &0.58&0.22&0.15     &0.62&0.22&-     &0.67&0.69&-     &\textbf{0.81$^*$}&\textbf{0.20$^*$}&\textbf{0.13} \\
    \textbf{ExpertQA}       &0.55&0.31&0.23     &0.57&0.28&-     &0.70&0.60&-     &\textbf{0.87$^*$}&\textbf{0.18$^*$}&\textbf{0.11$^*$} \\
    \midrule
    \textbf{DelucionQA}     &0.57&0.18&\textbf{0.10}     &\textbf{0.70$^*$}&0.22&-     &0.55&0.64&-     &0.64&\textbf{0.15$^*$}&\textbf{0.10} \\
    \textbf{EManual}        &0.54&0.17&\textbf{0.11$^*$}    &0.57&0.27&-     &0.61&0.64&-     &\textbf{0.76$^*$}&\textbf{0.13$^*$}&0.13 \\
    \textbf{TechQA}         &0.51&0.10&0.05     &0.52&0.12&-     &0.57&0.70&-     &\textbf{0.86$^*$}&\textbf{0.08$^*$}&\textbf{0.04$^*$} \\
    \midrule
    \textbf{FinQA}          &0.57&0.10&0.13     &0.57&\textbf{0.06$^*$}&-     &0.53&0.79&-     &\textbf{0.81$^*$}&0.10&\textbf{0.10} \\
    \textbf{TAT-QA}         &0.52&0.20&\textbf{0.17$^*$}     &0.63&\textbf{0.18$^*$}&-     &0.59&0.72&-     &\textbf{0.83$^*$}&0.27&0.23 \\
    \midrule
    \textbf{CUAD}           &0.51&0.27&0.11     &0.66&\textbf{0.19$^*$}&-     &0.40&0.66&-     &\textbf{0.80$^*$}&0.24&\textbf{0.10} \\

    \bottomrule
  \end{tabularx}
\end{table}
\endgroup

\section{Results}
\label{results}
Table \ref{table:main-results} reports results on test splits of each RAGBench component dataset. We compare baseline LLM methods with a finetunes DeBERTA encoder that trained on the full RAGBench train split.

% \paragraph{Current evaluators underperform on the RAG evaluation task} 
We observe that the finetuned DeBERTa model trained on RAGBench achieves competitive performance with billion-parameter LLM judges across numerous domain subsets of the RAGBench test set. On the hallucination detection task, DeBERTA AUROC scores range from 0.64 to 0.86. While RMSE for relevance and utilization range from 0.04 to 0.26, depending on the domain and task.

% We observe that the finetuned DeBERTa model outperforms the few/zero-shot LLM-judge baselines on most datasets. While GPT-3.5 demonstrates competitive performance with DeBERTa on a few metrics, DeBERTa consistently achieves superior performance metrics across all evaluations. Despite the versatility of LLM judges across various tasks, their lack of specialization necessitates finetuning for optimal results. Future work may focus on finetuning LLM judges to close the gap between DeBERTA and GPT-4 evaluation performance.

% In Appendix \ref{appendix:ood-deberta}, we demonstrate that, despite its small size, the finetuned DeBERTA model does generalize to out of domain RAG datasets in the same way that LLM-based approaches do.

% The LLM judges, though versatile on many tasks are not specialized enough. Finetuning is still required for optimal results. 

\paragraph{Estimating Context Relevance is Difficult} As shown in Table \ref{table:main-results}, Relevance RMSE scores are generally higher than those for Utilization, indicating a greater difficulty in the relevance prediction task. Utilization can be assessed through a straightforward semantic comparison between the context and the response. In contrast, relevance is a more intricate metric. Due to the nature of RAG, the majority of retrieved documents are semantically related to the query. However, mere semantic similarity is insufficient. The model must ascertain whether the provided context includes specific information necessary to accurately answer the question. Thus, the task inherently involves deriving the correct answer, followed by assessing what information in the context may be used to arrive at that answer.

% \paragraph{Performance degradation on long contexts}. In Figure \ref{figure-long-context-metrics}, we show how XX model performance degrades on long RAG context inputs. Explain how we got DeBERTA to work on long contexts. ONly show results up to 16k tokens. The broad range of context length included in RAGBench makes it a good dataset to use to improve RAG evaluation on long contexts.

\section{Conclusion}
\label{conclusion}
In this paper we introduce RAGBench, a large-scale dataset composed of real-world RAG examples intended for training and benchmarking powerful RAG \textit{evaluation models}. To this end, we formulate TRACe, a RAG evaluation framework comprising four metrics: u\textbf{T}ilization, \textbf{R}elevance, \textbf{A}dherence, and \textbf{C}ompleteness. TRACe standardizes the evaluation process, offering a consistent and systematic approach to measuring RAG system performance across various dimensions. We propose an automated approach to generate TRACe labels for RAGBench with an LLM and demonstrate high correlation between our approach and human judgements.

We benchmark existing RAG evaluation frameworks on RAGBench and demonstrate that a 400M parameter DeBERTa model finetuned on RAGBench data performs competitively with billion-parameter LLM Judges and commercial RAG evaluation systems. Though, the gap between the best-performing RAG evaluator and ground truth is still large. We motivate future work to leverage RAGBench toward fine-tuning more powerful evaluation models to explore the potential for narrowing the performance gap between these models and the ground truth.

Our contributions address the need for standardized benchmarks and methodologies, enabling more precise and actionable insights into the strengths and weaknesses of different RAG systems. This, in turn, will facilitate iterative improvement of RAG models, driving forward the capabilities of retrieval-augmented generation in real-world applications.

\bibliographystyle{abbrvnat}
\bibliography{RAGBench}

\section{Appendix}

\subsection{RAGBench Code and Data}
We release RAGBench data on Hugginggface: \url{https://huggingface.co/datasets/rungalileo/ragbench}. Refer to model card and documentation there.

We publish our inferfence and evaluation code on Gihub: \url{https://github.com/rungalileo/ragbench/tree/main/ragbench}.

\subsection{RAGBench Dataset Details}
\label{appendix-component-datasets}

RAGBench is sourced from publicly released acadmic and industry datasets. As far as we know, none of the component datasets contain personally identifiable information or offensive content.

\paragraph{PubMedQA \citep{jin-etal-2019-pubmedqa}} PubMedQA is a collection of PubMed research abstracts with corresponding yes/no/maybe questions paired with each abstract. The original dataset comprises 3 subsets: PQA-L, PQA-U, and PQA-A, with 1k, 60k, and 210k abstracts, respectively. For all subsets, the question is derived from the title of the PubMed article using rule-based heuristics. Long answers are automatically derived from the last sentence of the abstract for PQA-L and PQA-U, and QA-L answers are further reviewed by expert annotators and annotated as yes/no/maybe. PQA-A comprises exclusively automatically generated questions and short answers.

For RAGBench we utilize the PQA-U subset and re-frame it from QA into a RAG task. To simulate RAG, we leverage already segmented PQA-U abstracts context chunks and we encode them into a vector DB with OpenAI embeddings. The size of the resulting DB is 200k. We retrieve 4 chunks for each PQA-U question using FAISS with eucledian distance as the similarity function. We ignore the responses and labels in the original dataset and generate new responses with an LLM.

\paragraph{CovidQA-RAG} CovidQA-RAG is a combination of 2k expert-annotated questions sourced from COVID-QA \citep{moller-etal-2020-covidqa} and a vector database of 250,000 100-word passages built by \citet{siriwardhana-etal-2023-rag-for-odqa}. Both questions and answers are sourced from CORD-19 \citep{wang-etal-2020-cord} collection of research articles about COVID-19.

We embed the questions and database passages with OpenAI embeddings and retrieve up to N passages for each COVID-QA question from the vector database using FAISS with eucledian distance as the similarity function and max\_distance=0.25. We generate responses for each resulting RAG (context, question) instance with an LLM.

\paragraph{HotpotQA \citep{yang2018hotpotqa}} HotpotQA comprises 113K crowd-sourced question-answer pairs sourced from Wikipedia. Each pair is associated with a set of related context passages from one or multiple Wikipedia pages. The dataset is constructed in a way that requires multi-hop reasoning over multiple context documents to arrive at the answer, which renders it a valuable candidate for our benchmark. We sample data from the dev-distractor split, which contains up to 8 distractor context documents per sample. We downsample the context documents to 4 per example, making sure to include the document containing the response. We treat the context passages in HotpotQA as RAG context documents, and generate responses for each (context, question) instance with an LLM.

\paragraph{MS Marco \citep{nguyen2016msmacro}} MS Marco is an open-domain question answering dataset sourced from Bing search engine user query logs. Each question is associated with 10 context passages retrieved via Bing web search. Human annotators compose a response based on the provided context documents, and label the documents utilized in the response as relevant. We sample data from the original version of the dataset, comprising 80k train, 10k validation, and 10k test samples. As with other datasets, we ignore the human annotated answers and generate responses with an LLM in RAG setting.

\paragraph{CUAD \citep{hendrycks2021cuad}} CUAD is a collection of commercial legal contracts with expert annotated questions and responses. The contracts are sourced from a public legal contract library(EDGAR) and range from 1-100 pages in length. Experts in the legal domain compose multiple questions per contract and label the relevant parts of the contract that are useful for answering the questions. There are 21k questions pertaining to 510 documents in total. The questions are very specific to each contract, thus we don't perform additional retrieval over the contract corpus, and form RAG examples with 1 context contract each for our benchmark. Due to high anntoation costs associated with long-context RAG, we sample 5 question per doc. As with other datasets, we generate responses with an LLM in RAG setting.

\paragraph{DelucionQA \citep{sadat2024delucionqa}} DelucionQA is a domain-specific RAG dataset leveraging Jeep’s 2023 Gladiator model manual as the source of knowledge. The questions and answers are automatically generated by large language models. RAG context passages are retrieved from the Jeep car manual via both sparse and dense retrieval methods to add variance in the sample distribution. Further, MTurk workers annotate whether or not responses are supported by the context.

Upon closer inspection, we found only 1 relevant passage associated with each question in the DelucionQA dataset. To make the dataset more challenging for RAGBench, we build a vector database from the 1,046 context passages in DelucionQA and and retrieve up to 3 context documents per question from it. We use \texttt{text-embedding-ada-002} embeddings from OpenAI to build the database. There are 913 unique questions in DelucionQA. For each resulting (context, question) sample, we generate responses with an LLM.

% Since the dataset is already in RAG format, we do no additional retrieval or response generation. We adopt the same train/dev/test split as presented by the original authors.

\paragraph{EManual \citep{nandy-etal-2021-emanual}} EManual is a question answer dataset comprising consumer electronic device manuals and realistic questions about them composed by human annotators. The subset made available at the time of writing amounts to 659 unique questions about the Samsung Smart TV/remote and the accompanying user manual, segmented into 261 chunks. To form a RAG dataset, we embed the manual segments into a vector database with OpenAI embedding and retrieve up to 3 context documents per question from it. For each resulting (context, question) sample, we generate responses with an LLM.

\paragraph{TechQA \citep{castelli-etal-2020-techqa}} TechQA is a collection of real-world user questions posted on IBMDeveloper and DeveloperWorks forums, along with 50 technical support documents relating to each question. The documents are sourced from database of 800k technical documents that support accepted answers on the tech forums. The authors release 1.4k questions, split between train, validation, and test sets. The data are curated such that fractions on the each split unanswerable given the information in the linked documents, which makes it a good candidate for RAGBench. To reduce annotation costs, we sub-sample the data down to 10 documents per question, making sure to include the document containing the answer, when applicable.  We use the provided splits with (context document, question) examples and generate responses for each with an LLM.

% To form a RAG dataset, we embed the context database into a vector database with OpenAI embedding and retrieve up to Tk context documents per question from it. For each resulting (context, question) sample, we generate responses with an LLM.

% \paragraph{AttributionBench \citep{attributionbench}} AttributionBench is another curated RAG dataset sourced from various domains. We refer the reader to \citep{attributionbench} for a comprehensive  description of the AttributionBench component datasets and statistics. There are in total 13k training samples, 1k validation, and 1.6 in-domain and out-of-domain test samples, respectively. AttributionBench contains ground-truth context attribution labels as boolean attributed/not attributed value per each input claim and associated context documents. We leverage AttributionBench as a source of RAG (query, contexts, response) samples and assign new labels based on the CATCh framework defined in Section \ref{metric_definitions}.

\paragraph{FinQA \citep{chen-etal-2021-finqa}} FinQA is a QA dataset of financial report passages and associated questions. Questions are curated such that numerical reasoning over multiple unstructured and tabular inputs is required to arrive at the answer. FinQA totals 8,281 financial QA pairs, split between train, validation, and test splits. We retain the original splits and generate 2 LLM responses per each context-query example in FinQA.

\paragraph{TAT-QA \citep{zhu-etal-2021-tatqa}} TAT-QA is another financial QA dataset that requires numerical reasoning over tables and text. The data are sourced from 500 financial reports released on \url{https://www.annualreports.com/}. Expert annotators with background in finance annotate question-answer pairs based on the available documents. We leverage the full dataset (13k train, 1.6k validation and test) but generate new responses with LLMs for RAGBench.

\paragraph{HAGRID \citep{kamalloo2023hagrid}} HAGRID is a QA dataset built on top of MIRACL \citep{zhang2022making}, a multi-lingual information-retrieval dataset. HAGRID passes questions and relevant context documents from MIRACLE through an LLM to produce a response for each example in the dataset. Annotors then rate the response on informativeness and attribution dimensions. The original context documents are sourced from Wikipedia and associated questions are generated by expert annotators. Since HAGRID already contains LLM-generated responses, we directly use them and don't generate additional responses for RAGBench.

\paragraph{ExpertQA \citep{malaviya2024expertqa}} ExpertQA is a collection of curated questions from domain-experts in various fields of sicence, arts, and law. The dataset also contains expert curated passsages relevant to each question, alongside LLM-generated responses. As with HAGRID, we leverage the LLM-generated responses in ExpertQA directly for our RAG dataset.

\subsection{Response Generation Prompt}
\label{appendix-response-generation-prompt}
We use the following prompt template to generate LLM responses for each sample in RAGBench. Context documents, separated by line breaks, along with the question are slotted in for each generation sample.
\begin{verbatim}
    Use the following pieces of context to answer the question.

    {documents}

    Question: {question}
\end{verbatim}

\subsection{GPT Labeling Prompt}
\label{appendix-gpt-prompts}
We use the following prompt template to generate annotations with GPT-4
\begin{verbatim}
I asked someone to answer a question based on one or more documents.
Your task is to review their response and assess whether or not each sentence
in that response is supported by text in the documents. And if so, which 
sentences in the documents provide that support. You will also tell me which 
of the documents contain useful information for answering the question, and 
which of the documents the answer was sourced from.

Here are the documents, each of which is split into sentences. Alongside each
sentence is associated key, such as '0a.' or '0b.' that you can use to refer
to it:

```
{documents}
```

The question was:
```
{question}
```

Here is their response, split into sentences. Alongside each sentence is
associated key, such as 'a.' or 'b.' that you can use to refer to it. Note
that these keys are unique to the response, and are not related to the keys
in the documents:

```
{answer}
```

You must respond with a JSON object matching this schema:

```
{{
  "relevance_explanation": string,
  "all_relevant_sentence_keys": [string],
  "overall_supported_explanation": string,
  "overall_supported": boolean,
  "sentence_support_information": [
    {{
      "response_sentence_key": string,
      "explanation": string,
      "supporting_sentence_keys": [string],
      "fully_supported": boolean
    }},
  ],
  "all_utilized_sentence_keys": [string]
}}
```
The relevance_explanation field is a string explaining which documents 
contain useful information for answering the question. Provide a step-by-step
breakdown of information provided in the documents and how it is useful for
answering the question.

The all_relevant_sentence_keys field is a list of all document sentences keys
(e.g. '0a') that are revant to the question. Include every sentence that is
useful and relevant to the question, even if it was not used in the response,
or if only parts of the sentence are useful. Ignore the provided response when
making this judgement and base your judgement solely on the provided documents
and question. Omit sentences that, if removed from the document, would not 
impact someone's ability to answer the question.

The overall_supported_explanation field is a string explaining why the response
*as a whole* is or is not supported by the documents. In this field, provide a 
step-by-step breakdown of the claims made in the response and the support (or 
lack thereof) for those claims in the documents. Begin by assessing each claim
separately, one by one; don't make any remarks about the response as a whole 
until you have assessed all the claims in isolation.

The overall_supported field is a boolean indicating whether the response as a
whole is supported by the documents. This value should reflect the conclusion
you drew at the end of your step-by-step breakdown in overall_supported_explanation.

In the sentence_support_information field, provide information about the support
*for each sentence* in the response.

The sentence_support_information field is a list of objects, one for each sentence
in the response. Each object MUST have the following fields:
- response_sentence_key: a string identifying the sentence in the response.
This key is the same as the one used in the response above.
- explanation: a string explaining why the sentence is or is not supported by the
documents.
- supporting_sentence_keys: keys (e.g. '0a') of sentences from the documents that
support the response sentence. If the sentence is not supported, this list MUST
be empty. If the sentence is supported, this list MUST contain one or more keys.
In special cases where the sentence is supported, but not by any specific sentence,
you can use the string "supported_without_sentence" to indicate that the sentence
is generally supported by the documents. Consider cases where the sentence is 
expressing inability to answer the question due to lack of relevant information in
the provided contex as "supported_without_sentence". In cases where the sentence
is making a general statement (e.g. outlining the steps to produce an answer, or
summarizing previously stated sentences, or a transition sentence), use the 
sting "general".In cases where the sentence is correctly stating a well-known fact,
like a mathematical formula, use the string "well_known_fact". In cases where the
sentence is performing numerical reasoning (e.g. addition, multiplication), use 
the string "numerical_reasoning".
- fully_supported: a boolean indicating whether the sentence is fully supported by
the documents. 
  - This value should reflect the conclusion you drew at the end of your step-by-step
  breakdown in explanation. 
  - If supporting_sentence_keys is an empty list, then fully_supported must be false. 
  - Otherwise, use fully_supported to clarify whether everything in the response 
  sentence is fully supported by the document text indicated in supporting_sentence_keys
  (fully_supported = true), or whether the sentence is only partially or incompletely
  supported by that document text (fully_supported = false).

The all_utilized_sentence_keys field is a list of all sentences keys (e.g. '0a') that
were used to construct the answer. Include every sentence that either directly supported
the answer, or was implicitly used to construct the answer, even if it was not used 
in its entirety. Omit sentences that were not used, and could have been removed from
the documents without affecting the answer.

You must respond with a valid JSON string.  Use escapes for quotes, e.g. `\\"`, and 
newlines, e.g. `\\n`. Do not write anything before or after the JSON string. Do not 
wrap the JSON string in backticks like ``` or ```json.

As a reminder: your task is to review the response and assess which documents contain
useful information pertaining to the question, and how each sentence in the response
is supported by the text in the documents.\
\end{verbatim}

\subsection{Annotation Post-Processing Steps}
\label{appendix:annotation-post-processing}
As shown in Appendix \ref{appendix-gpt-prompts}, we request very detailed annotations with explanations from GPT-4-turbo. We pivot on chain-of-thought \citep{wei-2022-chain-of-thought} and redundancy to encourage high quality labels from the annotator model.

For Adherence, we request both response-level and sentence-level annotations that we compare in post-processing to identify inconsistencies where GPT-4 disagrees with its own judgements. For example, if GPT-4 claims a response as supported by the context as a whole, but identifies no supporting information for one or more claims in the response, we send the example for re-annotation. We re-annotate all data up to 3 times, after which a fraction (<2\%) of the data are still conflicting. After manual inspection, we find that the majority of the conflicts arise from partially hallucinated sentences that are somewhat, but not fully, grounded in the context. We leverage a sentence-level "fully\_supported" boolean annotation to identify and resolve such cases. According to our annotation schema, we treat all partially supported sentences as hallucinations.

Since all TRACe metrics are related, we qualitatively observe that taking the extra measures for Adherence also positively impacts the quality and stability of the relevance and utilization labels. 

In the final post-processing step, we remove any off-schema keys that GPT-4-turbo sometimes injects into the response. For example, it will occasionally misspell "supporting\_sentence\_keys" as "support\textbf{ed}\_sentence\_keys" and/or introduce completely new fields into the output json. We algorithmically find and remove/replace such annotation errors.

\subsection{Annotation Alignment with Human Judgements}
\label{appendix-annotation-alignment}
\subsubsection{Adherence Alignment with DelucionQA}
We validate our metric formulations and labeling approach on a human annotated benchmark. DelucionQA \citep{sadat2024delucionqa} is a curated collection of user queries on the operation of Jeep’s 2023 Gladiator model. Natural language queries are first generated by an LLM, then reviewed and filtered by human annotators. Context documents are sourced from Jeep’s Gladiator User Manual, and responses are generated by various LLMs. Human annotators label each response sentence as "Supported" by the context documents, "Conflicted", or "Neither". Example-level labels are derived from the span-level annotation as follows: if at least one response sentence is annotated as "Conflicted" or "Neither", the whole response receives a Hallucinated label. 

In our initial investigation, we found that sentences that DelucionQA labels as "Neither" often fall into one of two categories: (1) general filler statements (e.g. "Here are the steps:"), (2) claims of missing information (e.g. "There is no mention of any problem with engine start-up in freezing weather related to DEF."). According to our annotation schema, these types of statements are generally grounded in the context and not hallucinations. Thus, we remove examples with any "Neither" sentence annotations for our analysis. We annotate the remaining 421 examples with our LLM annotator and report alignment with human annotations in Table \ref{table:annotation-validation-human-alignment}.

\begingroup
% \newcolumntype{s}{>{\hsize=.5\hsize}X}
\renewcommand{\arraystretch}{1.2} % increase vertical spacing between rows
\begin{table}
  \caption{Annotation Alignment with DelucionQA. We report F1 and Accuracy metrics on human annotated subsets of DelucionQA. \textbf{DelucionQA(40)} is a subset of 40 examples randomly sampled from the DelucionQA test set and annotated for Relevance and Utilization by the authors.}
  \label{table:annotation-validation-human-alignment}
  \centering
  \small
  \begin{tabularx}{0.8\textwidth}{lccc}
    \toprule
     Test Set & Metric & F1 & Accuracy \\
    \midrule
    DelucionQA - example level & Adherence & 0.83 & 0.76 \\
    DelucionQA - example level - remove "Neither" & Adherence & 0.96 & 0.93 \\
    DelucionQA - span level - remove "Neither" & Adherence & 0.97 & 0.95 \\
    DelucionQA(40) - sapen level & Utilization & 0.92 & 0.94 \\
    DelucionQA(40) - span level & Relevance & 0.76 & 0.78 \\
    \bottomrule
  \end{tabularx}
\end{table}
\endgroup

\subsubsection{Relevance and Utilization Alignment with DelucionQA}
To validate Relevance and Utilization annotations, the authors annotate a small set of 40 randomly samples examples from the DelucionQA test set. We follow the same instructions as in out annotation prompt \ref{appendix-gpt-prompts} to label relevant and utilized context sentences, given the context, query, and response. We report sentence-level F1 and overall alignment (Accuracy) scores in Table \ref{table:annotation-validation-human-alignment}.

\subsubsection{Rank-based Alignment for Adherence and Relevance}
We use mock RAG datasets generated by \citet{saadfalcon2024ares} for this analysis. Their RAG validation set is sampled from KILT \citep{petroni-etal-2021-kilt}, including Natural Questions (NQ)\citep{natural-questions}, HotpotQA\citep{yang2018hotpotqa}, FEVER\citep{thorne-etal-2018-fever}, and Wizards of Wikipedia (WoW) \citep{dinan2019wizard} datasets. The authors synthetically generate systems of varying quality by adjusting the ratio of relevant documents and responses in the data. We sample 500 examples from each simulated RAG dataset and annotated them as described in section \ref{section:llm-annotator}. Next, we calculate average annotated context relevance and adherence scores for each dataset and use those to rank the mock systems. We compare our rankings to ground truth with the Kendall rank correlation (Kendall's $\tau$) metric, which evaluates the agreement between two sets of ranks on a scale from 0 (no agreement) to 1 (perfect agreement).

As shown in Table \ref{table:annotation-validation}, the GPT-4 annotations achieve high Kendall's $\tau$ ranging from 0.78 to 1. For a fair comparison with the ground truth labels, we derive binary context relevance and labels from the GPT-4 annotations by thresholding the example Relevance score (equation \ref{eq:example-relevance}) at 0. For comparison, we also report ranking results with our more granular example-level Relevance scores that range from 0-1. We find that these metrics produce a different ranking (see lower Kendall's $\tau$ in Table \ref{table:annotation-validation}), which we attribute to the metrics capturing differences in retrieved context length across the different examples.

\begingroup
% \newcolumntype{s}{>{\hsize=.5\hsize}X}
\renewcommand{\arraystretch}{1.2} % increase vertical spacing between rows
\begin{table}
  \caption{\textbf{Ranking of Simulated RAG Systems.} We evaluate GPT-4-turbo annotations on simulated RAG datasets from \citet{saadfalcon2024ares}. The data from each source are synthetically augmented to create sets with increasing degrees of \textbf{context relevance} (Rel) and \textbf{answer adherence} (Adh). We annotate 500 samples from each set and rank them according to the average \textbf{context relevance} and \textbf{answer adherence} metrics. We report Kendall's tau to evaluate the agreement between GPT-4-turbo rankings and ground truth (higher is better).}
  \label{table:annotation-validation}
  \centering
  \small
  \begin{tabularx}{0.8\textwidth}{lcccccccc}
    \toprule
     & \multicolumn{2}{c}{NQ} & \multicolumn{2}{c}{HotpotQA} & \multicolumn{2}{c}{WoW} & \multicolumn{2}{c}{FEVER}  \\
    \cmidrule(r){2-3}   \cmidrule(lr){4-5} \cmidrule(lr){6-7} \cmidrule(lr){8-9}
     & Rel & Adh & Rel & Adh & Rel & Adh & Rel & Adh \\
    \midrule
    Kendall's Tau binary  & 1.0 & 0.83 % NQ
                       & 0.87 & 1.0 % Hotpot
                       & 1.0 & 0.89 % Wow
                       & 1.0 & 0.78 \\% FEVER
    Kendall's Tau continuous & 0.94 & - % NQ
                        & 0.73 & - % Hotpot
                        & 1.0 & - % Wow
                        & 0.77 & - \\% FEVER
    \bottomrule
  \end{tabularx}
\end{table}
\endgroup

\subsection{RAG Case Study}
\label{appendix-rag-case-study}

We use the following prompt templates for the RAG Case Study:

NO CONTEXT:
\begin{verbatim}
{question}
\end{verbatim}

SHORT:
\begin{verbatim}
Answer the question using the provided context.

Context:
{documents}

Question: {question}
\end{verbatim}

LONG:
\begin{verbatim}
You are a chatbot providing answers to user queries. You will be given one or more context documents, and a question. \
Use the information in the documents to answer the question.

If the documents do not provide enough information for you to answer the question, then say \
"The documents are missing some of the information required to answer the question." Don't quote any external knowledge that is \
not in the documents. Don't try to make up an answer.

Context Documents:
{documents}

Question: {question}
\end{verbatim}

LONG + CoT:
\begin{verbatim}
You are a chatbot providing answers to user queries. You will be given one or more context documents, and a question. \
Use the information in the documents to answer the question.

If the documents do not provide enough information for you to answer the question, then say \
"The documents are missing some of the information required to answer the question." Don't quote any external knowledge that is \
not in the documents. Don't try to make up an answer.

Think step by step and explain your reasoning, quoting the documents when necessary.

Context Documents:
{documents}

Question: {question}
\end{verbatim}
 
Table \ref{table:rag-case-study-full-results} reports comprehensive results from the RAG Case Study.

\begingroup
% \newcolumntype{s}{>{\hsize=.5\hsize}X}
\renewcommand{\arraystretch}{1.2} % increase vertical spacing between rows
\begin{table}
  \caption{RAG Case Study Results.}
  \label{table:rag-case-study-full-results}
  \centering
  \tiny
  \begin{tabularx}{\textwidth}{lrllrrrr}
    \toprule
     retriever & k & prompt version & generation model & ave relevance & ave utilization & ave completeness & pc hallucinated \\
        \midrule
        tfidf & 2 & no context & gpt-3.5-turbo-0125 & 0.65 & 0.30 & 0.53 & 0.94 \\
        tfidf & 2 & no context & gpt-4o & 0.66 & 0.45 & 0.70 & 0.95 \\
        tfidf & 2 & short & gpt-3.5-turbo-0125 & 0.62 & 0.43 & 0.72 & 0.19 \\
        tfidf & 2 & short & gpt-4o & 0.66 & 0.54 & 0.81 & 0.29 \\
        tfidf & 2 & long & gpt-3.5-turbo-0125 & 0.62 & 0.43 & 0.71 & 0.04 \\
        tfidf & 2 & long & gpt-4o & 0.66 & 0.35 & 0.57 & 0.08 \\
        tfidf & 2 & long + CoT & gpt-3.5-turbo-0125 & 0.65 & 0.53 & 0.82 & 0.13 \\
        tfidf & 2 & long + CoT & gpt-4o & 0.63 & 0.50 & 0.78 & 0.04 \\
        tfidf & 4 & no context & gpt-3.5-turbo-0125 & 0.46 & 0.22 & 0.53 & 0.86 \\
        tfidf & 4 & no context & gpt-4o & 0.45 & 0.29 & 0.65 & 0.86 \\
        tfidf & 4 & short & gpt-3.5-turbo-0125 & 0.43 & 0.29 & 0.67 & 0.18 \\
        tfidf & 4 & short & gpt-4o & 0.47 & 0.38 & 0.78 & 0.13 \\
        tfidf & 4 & long & gpt-3.5-turbo-0125 & 0.45 & 0.28 & 0.64 & 0.05 \\
        tfidf & 4 & long & gpt-4o & 0.48 & 0.28 & 0.60 & 0.05 \\
        tfidf & 4 & long + CoT & gpt-3.5-turbo-0125 & 0.44 & 0.33 & 0.72 & 0.10 \\
        tfidf & 4 & long + CoT & gpt-4o & 0.49 & 0.36 & 0.73 & 0.02 \\
        faiss & 2 & no context & gpt-3.5-turbo-0125 & 0.81 & 0.40 & 0.50 & 0.94 \\
        faiss & 2 & no context & gpt-4o & 0.81 & 0.53 & 0.66 & 0.87 \\
        faiss & 2 & short & gpt-3.5-turbo-0125 & 0.75 & 0.55 & 0.72 & 0.07 \\
        faiss & 2 & short & gpt-4o & 0.82 & 0.72 & 0.86 & 0.11 \\
        faiss & 2 & long & gpt-3.5-turbo-0125 & 0.78 & 0.58 & 0.71 & 0.04 \\
        faiss & 2 & long & gpt-4o & 0.82 & 0.52 & 0.62 & 0.10 \\
        faiss & 2 & long + CoT & gpt-3.5-turbo-0125 & 0.81 & 0.64 & 0.76 & 0.07 \\
        faiss & 2 & long + CoT & gpt-4o & 0.83 & 0.69 & 0.82 & 0.04 \\
        faiss & 4 & no context & gpt-3.5-turbo-0125 & 0.58 & 0.27 & 0.47 & 0.88 \\
        faiss & 4 & no context & gpt-4o & 0.58 & 0.36 & 0.60 & 0.79 \\
        faiss & 4 & short & gpt-3.5-turbo-0125 & 0.53 & 0.31 & 0.59 & 0.14 \\
        faiss & 4 & short & gpt-4o & 0.58 & 0.47 & 0.78 & 0.07 \\
        faiss & 4 & long & gpt-3.5-turbo-0125 & 0.54 & 0.33 & 0.61 & 0.08 \\
        faiss & 4 & long & gpt-4o & 0.61 & 0.32 & 0.52 & 0.09 \\
        faiss & 4 & long + CoT & gpt-3.5-turbo-0125 & 0.58 & 0.42 & 0.71 & 0.09 \\
        faiss & 4 & long + CoT & gpt-4o & 0.60 & 0.47 & 0.76 & 0.05 \\
    \bottomrule
  \end{tabularx}
\end{table}
\endgroup

\subsection{DeBERTa model training}
\label{appendix-deberta-training}

We train the model on a Google Cloud Platform A-100 GPU instance for 3 epochs with initial learning rate $5^{-6}$ for the base model layers and $2^{-5}$ for the heads, with warmup and a linear decay rate.

\end{document}